\documentclass{article}

\usepackage[english]{babel}
\usepackage{listings}
\usepackage{xcolor}

\usepackage[letterpaper,top=2cm,bottom=2cm,left=3cm,right=3cm,marginparwidth=1.75cm]{geometry}

\usepackage{amsmath}
\usepackage{graphicx}
\usepackage[colorlinks=true, allcolors=blue]{hyperref}

\title{Off-policy Evaluation for Payments at Adyen}
\author{Alex Egg \\ Adyen \\ \texttt{alex.egg@adyen.com}}
\date{}

\begin{document}
\maketitle

\begin{abstract}
This paper demonstrates the successful application of Off-Policy Evaluation (OPE) to accelerate recommender system development and optimization at Adyen, a global leader in financial payment processing. Facing the limitations of traditional A/B testing, which proved slow, costly, and often inconclusive, we integrated OPE to enable rapid evaluation of new recommender system variants using historical data. Our analysis, conducted on a million-scale dataset of transactions, reveals a strong correlation between OPE estimates and online A/B test results, projecting an incremental 9--54 million transactions over a six-month period. We delve into the practical challenges and trade-offs associated with deploying OPE in a high-volume production environment, including leveraging exploration traffic for data collection, mitigating variance in importance sampling, and ensuring scalability through the use of Apache Spark. By benchmarking various OPE estimators, we provide guidance on their effective utilization and integration into the decision-making process for financial payment systems. 
\end{abstract}

\section{Introduction}

Recommender systems have become ubiquitous across industry, driving personalized experiences and optimizing performance in decision making systems. At Adyen, a global payments provider, recommender systems play a crucial role in maximizing transaction authorization rates. These systems operate by suggesting interventions at the time of payment, such as adjusting logic or modifying transaction metadata, to improve the likelihood of successful authorization. This optimization is vital due to technical integration inconsistencies between scheme specifications and the respective issuing (or otherwise) bank implementations, creating opportunities to enhance authorization rates through targeted interventions. However, designing, implementing and especially testing these decision making policies is a notoriously complex challenge.

Traditionally, A/B testing, or On-policy Evaluation, has been the primary method for evaluating decision making policies. However, this approach suffers from significant drawbacks: it is time-consuming, incurs high opportunity costs, and limits the speed of iteration. One analysis of A/B testing practices at Adyen revealed that 58\% of tests were flat or inconclusive, resulting in over 20 weeks per year of wasted experimentation time. This inefficiency hinders innovation and delays the deployment of impactful changes.

To address these challenges, we turned to Off-Policy Evaluation (OPE)---a powerful technique that allows for the evaluation of new models or policies by using historical data. By leveraging OPE, we aimed to achieve faster decision-making, reduce reliance on costly A/B tests, and accelerate the identification of winning variants. This paper details our experience implementing and evaluating OPE at Adyen, focusing on the practical challenges, key findings, and resulting impact on our recommender system optimization process.

While OPE has shown promise in academic research, its application to large-scale industrial recommender systems remains relatively unexplored. Existing OPE literature often focuses on smaller datasets and controlled environments, leaving open questions about its effectiveness and scalability in real-world, high-volume settings. This paper directly addresses this gap by demonstrating the successful implementation and evaluation of OPE on million-scale datasets within Adyen's production environment. Our work provides valuable insights and practical guidance for applying OPE to large-scale recommender systems, paving the way for wider adoption in the industry.

This paper is structured as follows: Section~\ref{sec:related-work} summarizes relevant work on OPE in industry contexts. Section~\ref{sec:background} provides a background on OPE estimators. Section~\ref{sec:method} details the methodology for implementing OPE at Adyen, including the challenges and solutions in the industrial setting. Section~\ref{sec:experiment} describes the experimental setup. Section~\ref{sec:results} presents the results and analysis of our OPE implementation. Section~\ref{sec:discussion} discusses the broader implications and limitations of our findings. Finally, Section~\ref{sec:conclusion} concludes the paper and suggests directions for future work.

\section{Related Work}
\label{sec:related-work}

Off-policy evaluation (OPE) has become an active research area, driven partly by the need to reduce the cost and risk of online experimentation. Several large-scale industrial and applied studies demonstrate the potential of OPE, particularly in recommender systems and advertising domains:

\begin{itemize}
    \item \textbf{News and Content Recommendation.}  
    Li et al.~\cite{li2011unbiased} showed how Inverse Propensity Scoring (IPS) can be leveraged for unbiased offline evaluation of contextual-bandit-based news article recommendations, reducing the need for prolonged online A/B testing.  
    \item \textbf{E-commerce and Advertising.}  
    Criteo, an online advertising platform, has explored IPS and related importance-sampling estimators to evaluate ad personalization models using historical click logs~\cite{bottou2013counterfactual}. Similarly, Alibaba and JD.com have reported success in using doubly robust estimators to quickly iterate on recommender system updates~\cite{zhou2018offline}.
    \item \textbf{Video Streaming and Media.}  
    Netflix has publicly documented counterfactual evaluation techniques to reduce experimentation overhead in personalized movie recommendations~\cite{gomez2021netflix}. Their large user base and diversified content catalog pose similar challenges to our high-volume financial setting, such as high variance and reward sparsity.  
    \item \textbf{Multi-World Testing and Open-Source Tools.}  
    Microsoft’s Multi-World Testing (MWT) platform~\cite{agarwal2016making} supports large-scale deployment of contextual bandit algorithms and features robust OPE libraries (e.g., IPS, SNIPS, DR). This has allowed systematic offline evaluation of new ranking policies before minimal online validation.  
    \item \textbf{Open Bandit Pipeline.}  
    Zolna et al.~\cite{tholz2021open} introduced an open-source pipeline for evaluating and benchmarking OPE methods on real-world bandit data, emphasizing reliability and reproducibility.
\end{itemize}

These large-scale applications, particularly those by Alibaba, JD.com, Netflix, Microsoft, and Criteo, underscore the feasibility of OPE in production settings. They also highlight common pitfalls, such as high variance, data sampling bias, and the importance of careful exploration strategy design. Our work at Adyen extends this body of knowledge by applying OPE to large-scale financial payment data.

\section{Background}
\label{sec:background}

Off-Policy Evaluation (OPE) aims to estimate the performance of a new policy (the \textit{target policy}) using data collected under a different policy (the \textit{logging policy}). This is crucial in recommender systems where deploying a new policy directly for evaluation (e.g., through A/B testing) can be costly and time-consuming. OPE leverages historical data to provide insights into the potential effectiveness of new policies \textit{before} they are launched.

Several OPE estimators have been proposed, each with its own strengths and weaknesses. Here, we provide an overview of four key estimators: Direct Method (DM), Inverse Propensity Scoring (IPS), Self-Normalized Importance Sampling (SNIPS), and Doubly Robust (DR).

\subsection{Direct Method (DM)}

The Direct Method (DM) is a model-based approach that directly estimates the expected reward of the target policy by learning a regression model to predict rewards from context and action features~\cite{li2010contextual}:

\begin{equation*}
    \hat{V}_{DM}(\pi_t) = \frac{1}{n} \sum_{i=1}^{n} \hat{r}(x_i, \pi_t(x_i)),
\end{equation*}

where

\begin{itemize}
    \item $\hat{V}_{DM}(\pi_t)$ is the estimated value of the target policy $\pi_t$.
    \item $n$ is the number of logged interactions.
    \item $\hat{r}(x_i, \pi_t(x_i))$ is the predicted reward for context $x_i$ and action $\pi_t(x_i)$ chosen by the target policy.
\end{itemize}

While DM is straightforward to implement and often exhibits low variance, it is highly susceptible to bias if the reward model is misspecified.

\subsection{Inverse Propensity Scoring (IPS)}

Inverse Propensity Scoring (IPS) is a model-free approach that reweights the observed rewards in the logged data to account for the difference between the logging and target policies. It utilizes importance sampling to correct for the distribution shift:

\begin{equation*}
    \hat{V}_{IPS}(\pi_t) = \frac{1}{n} \sum_{i=1}^{n} \frac{\pi_t(a_i | x_i)}{\pi_o(a_i | x_i)} r_i,
\end{equation*}

where

\begin{itemize}
    \item $\hat{V}_{IPS}(\pi_t)$ is the estimated value of the target policy $\pi_t$.
    \item $\pi_t(a_i \mid x_i)$ is the probability of the target policy taking action $a_i$ in context $x_i$.
    \item $\pi_o(a_i \mid x_i)$ is the probability of the logging policy taking action $a_i$ in context $x_i$.
    \item $r_i$ is the observed reward for the $i$-th interaction.
\end{itemize}

IPS provides unbiased estimates when the logging policy has sufficient \textit{support} (i.e., non-zero probability for all actions that the target policy might take). However, it can suffer from high variance, especially when the polices are divergent or importance weights are large or when data is limited.

\textit{The following two methods attempt to control the variance of IPS in different ways by use of \textbf{Control Variates}.}

\subsection{Self-Normalized Importance Sampling (SNIPS)}

Self-Normalized Importance Sampling (SNIPS) \cite{swaminathan2015self} is a variant of IPS that addresses the high variance issue by normalizing the importance weights. This normalization helps to reduce the influence of extreme weights and improves the stability of the estimator.

\begin{equation*}
    \hat{V}_{SNIPS}(\pi_t) = \frac{\sum_{i=1}^{n} \frac{\pi_t(a_i | x_i)}{\pi_o(a_i | x_i)} r_i}{\sum_{i=1}^{n} \frac{\pi_t(a_i | x_i)}{\pi_o(a_i | x_i)}} 
\end{equation*}

SNIPS is consistent but can be biased. However, it often performs well empirically, particularly in low-data regimes, and does not require hyperparameter tuning. The advantages of  the SNIPS estimator come with the trade-off that the algorithm learning is no-longer out-of-core.

\subsection{Doubly Robust (DR)}

The Doubly Robust (DR) estimator \cite{dudik2014doubly,thomas2016data} combines the strengths of both DM and IPS. It leverages a reward model to reduce variance while using importance sampling to correct for bias.

\begin{equation*}
    \hat{V}_{DR}(\pi_t) = \frac{1}{n} \sum_{i=1}^{n} \left[ \frac{\pi_t(a_i | x_i)}{\pi_o(a_i | x_i)} (r_i - \hat{r}(x_i, a_i)) + \frac{1}{n} \sum_{i=1}^{n}  \hat{r}(x_i, \pi_t(x_i)) \right]
\end{equation*}

DR is more robust than either DM or IPS alone. It is less sensitive to model misspecification than DM, less affected by high variance than IPS and is unbiased if either $\hat{r}(x,y) = r(x,y)$ or $\hat{p_i} = \pi_0(y_i | x_i)$ .

\section{Methodology}
\label{sec:method}

Due to our billion-scale dataset we implemented and evaluated four key OPE estimators: DM, IPS, SNIPS, and DR in Spark. A minimal concrete example of the implementation is shown below.

\subsection{Implementations}

% Define style for code
\lstset{
    language=Python,
    basicstyle=\ttfamily\footnotesize,
    keywordstyle=\color{blue},
    commentstyle=\color{green},
    stringstyle=\color{red},
    frame=single,
    numbers=left,
    numberstyle=\tiny\color{gray},
    breaklines=true,
    showstringspaces=false
}

% Include your code in a section
\begin{lstlisting}[caption={Example implementation of IPS Estimation using Spark and a custom IPSEstimator.}, label={code:ips_example}, captionpos=b]
from lib_ope.estimators import IPSEstimator
from pyspark.sql import SparkSession

# Initialize Spark Session
spark = SparkSession.builder.appName("OPE Example").getOrCreate()

# Sample DataFrame with 'metric', 'p_pi', and 'p_0' columns
df = spark.createDataFrame([
    (1.0, 0.8, 0.6),
    (0.5, 0.7, 0.5),
    (1.5, 0.9, 0.7)
], ["metric", "p_pi", "p_0"])

# Build IPS estimator
ips_estimator = IPSEstimator(df.metric, df.p_pi, df.p_0)
ips_estimate = ips_estimator.estimate()

# Apply the estimator across logged feedback
df = df.withColumn("ips_estimate", ips_estimate)
results = df.select("ips_estimate").collect()
print(results)
\end{lstlisting}
The following implementations can be viewed in the \textit{model the world} vs \textit{model the bias} framework as Value-based and Model-free implementations respectively. 

\subsubsection{Value-based}
Value-based methods attempt to model the world and build a reward model to directly estimate the metric of choice.

\textbf{Direct Method.} In the bias/variance tradeoff, bias is introduced as a product of modeling the complicated hidden reward function which, depending on the task, will be biased.

One must collect a dataset from the logging policy $\pi_0$ with $S = (x_i,y_i,\delta_i)$ where $i$ is a time step, so $x_i$ means all the state of the system at $i$.

We trained a regression model using gradient boosting to predict rewards based on context and action features. This model was then used to \textit{directly} estimate the performance of target policies on the logged data.

This model gets around 80\% accuracy in ab tests.

\subsubsection{Model-free}

Model-free approaches attempt to model the bias instead of the world. Typically they are based off of some type of reweighing scheme where the observed reward $\delta_i$ from the logging policy is reweighted by ratio of the probably of the action being taken by the target and logging policy respectively $\frac{\pi_t(a_i | x_i)}{\pi_o(a_i | x_i)}$.

There is an important and subtle implication in the reweighting scheme: it implies that the logging policy is stochastic or in other words you have a probability distribution across actions. This is important to highlight in light of the fact that most industrial applications will be deterministic by default, for example following some binomial objective click objective $P(C | ctx, action)$. In this case we don't have a distribution over actions but rather a probability of conversion for example. This is also the case at Adyen where the existing models are based on a binomial objective in a epsilon-greedy policy. \cite{vangara2024contextual}

To overcome this issue, the lack of action probability in the logs, we can exploit the fact that we are in an epsilon-greedy framework and that for the exploration traffic we have implied action probabilities. Using this intuition we can recover a stochastic logging policy which puts us in the reweighting framework. However, this comes w/ the tradeoff that the target and logging polices will be very \textit{divergent}. Divergent polices will experience large weights which will potentially contribute to variance that we will have to mitigate. Ideally our logging policy would be stochastic which would result in smaller weights and less variance and better OPE estimate. -- this is something our team are prioritizing for the future.

\textbf{Inverse Propensity Scoring.} We implemented IPS using the importance weights derived from the exploration traffic. We applied variance reduction techniques, such as weight clipping and baseline models, to improve the stability of the estimates.

\begin{lstlisting}[caption={Spark implementation of the IPSEstimator class for OPE.}, label={code:ips_estimator}, captionpos=b]
class IPSEstimator(OPEEstimator):
    def __init__(self, reward_col: Column, target_propensity: Column, logging_propensity: Column):
        """
        Parameters:
            reward_col (Column): Column with reward values.
            target_propensity (Column): Column with target propensity values.
            logging_propensity (Column): Column with logging propensity values.
        """
        self.reward_col = reward_col
        self.target_propensity = target_propensity
        self.logging_propensity = logging_propensity

    def estimate(self) -> Column:
        weight: Column = self.target_propensity / self.logging_propensity
        ips_estimate = self.reward_col * weight
        return ips_estimate
\end{lstlisting}

\textbf{Self-Normalized Importance Sampling (SNIPS):} We also implemented SNIPS as an alternative to IPS, which offers improved stability by normalizing the importance weights using the expected sample size as multiplicative control variate.

\begin{lstlisting}[caption={Spark implementation of the SNIPSEstimator class, extending IPSEstimator for normalized importance sampling.}, label={code:snips_estimator}, captionpos=b]
class SNIPSEstimator(IPSEstimator):
    def __init__(self, reward_col: Column, target_propensity: Column, logging_propensity: Column, E_weight: Column):
        """
        Parameters:
            reward_col (Column): Column with reward values.
            target_propensity (Column): Column with target propensity values.
            logging_propensity (Column): Column with logging propensity values.
            E_weight (Column): Column with expected weights.
        """
        super().__init__(reward_col, target_propensity, logging_propensity)
        self.E_weight = E_weight

    def estimate(self) -> Column:
        ips_estimate = super().estimate()
        snips = ips_estimate / self.E_weight
        return snips
\end{lstlisting}

\textbf{Doubly Robust (DR):} We combined DM and IPS to leverage their respective strengths and create a more robust estimator. DR mimics IPS to use a weighted version of rewards, but DR also uses the estimated mean reward function (the regression model) as a control variate to decrease the variance.

\begin{lstlisting}[caption={Implementation of the DREstimator class, extending IPSEstimator for Doubly Robust (DR) estimation.}, label={code:dr_estimator}, captionpos=b]
class DREstimator(IPSEstimator):
    def __init__(
        self,
        reward_col: Column,
        target_propensity: Column,
        logging_propensity: Column,
        DM_estimate: Column,
        E_DM: Column,
    ):
        """
        Parameters:
            reward_col (Column): Column with reward values.
            target_propensity (Column): Column with target propensity values.
            logging_propensity (Column): Column with logging propensity values.
            DM_estimate (Column): Column with direct method estimate.
            E_DM (Column): Column with expected direct method estimate (control variate).
        """
        super().__init__(reward_col, target_propensity, logging_propensity)
        self.DM = DM_estimate
        self.E_DM = E_DM

    def estimate(self) -> Column:
        weight: Column = self.target_propensity / self.logging_propensity
        dr = weight * (self.reward_col - self.DM) + self.E_DM
        return dr
\end{lstlisting}

\section{Experiment}
\label{sec:experiment}

To benchmark the OPE estimators, we evaluated their performance by comparing their estimates to actual on-policy (A/B test) results. For each completed A/B test, we computed both the on-policy metric estimate and the corresponding OPE estimate using logged data. This process was repeated across numerous A/B tests conducted over a year, allowing us to assess the correlation and accuracy of OPE estimators under real-world conditions.

Our dataset included billions of interactions logged from Adyen's recommender systems, including context features for the underlying model and the action and restive reward observed from the issuer bank (environment). 

Evaluation metrics included Pearson Correlation to measure agreement between OPE estimates and online metrics, and Root Mean Square Error (RMSE) to quantify the accuracy of OPE estimates. Baseline methods included naive estimators and results from traditional A/B tests.

\section{Results}
\label{sec:results}

\begin{figure}[h]
    \centering
    \includegraphics[width=\textwidth]{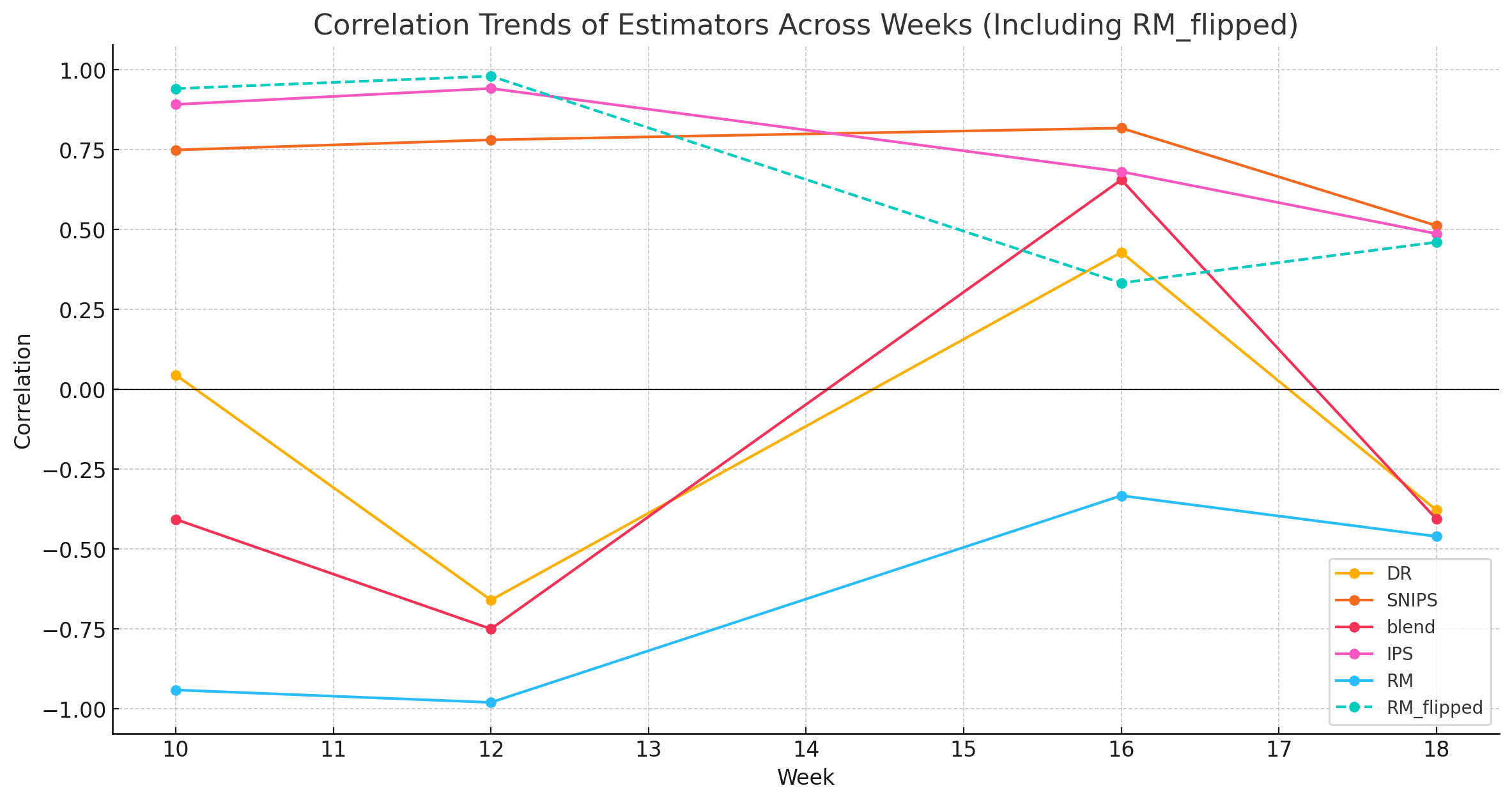}
    \caption{Correlation Trends of Estimators Across Weeks. Note: RM means "Reward Model" or the Direct Method as described in the literature. Blend is the average of all the estimators.}
    \label{fig:correlation_trends}
\end{figure}

Our experiments revealed that, among the OPE estimators, IPS and SNIPS consistently demonstrated strong correlation with online A/B test results, while DM and DR exhibited weaker performance. Figure \ref{fig:correlation_trends} shows the weekly Pearson correlation coefficients between OPE estimates and online metrics. Notably, IPS and SNIPS consistently maintain a correlation above 0.8 across all weeks, indicating their robust ability to predict the performance of new recommender system variants using historical data.

\begin{figure}[h]
    \centering
    \includegraphics[width=\textwidth]{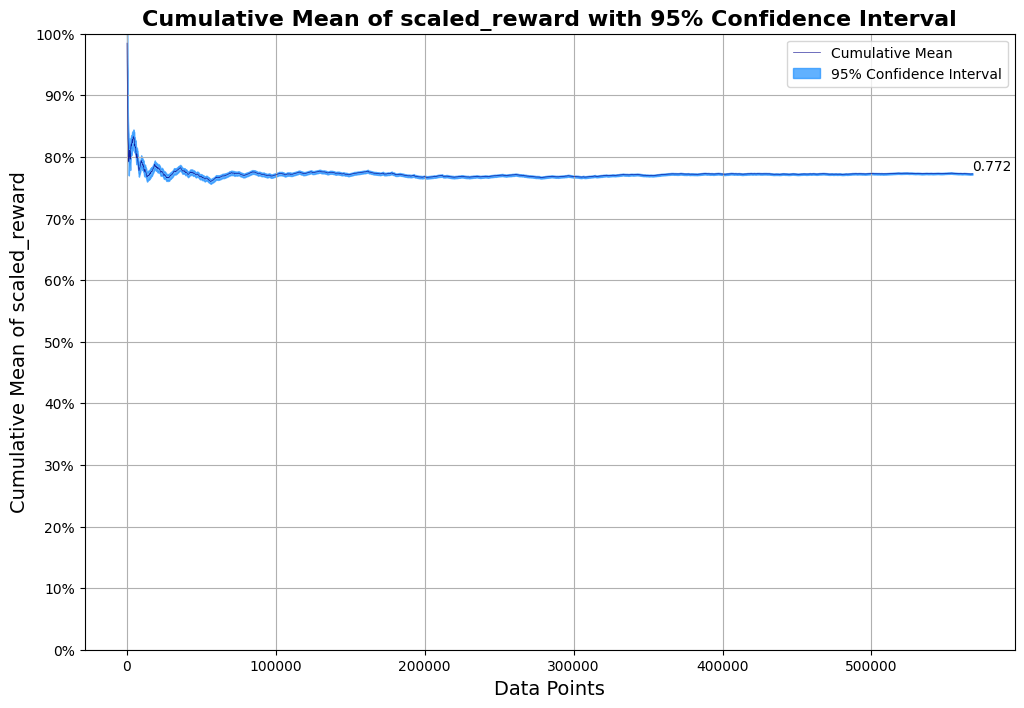} 
    \caption{Variance of IPS Estimator across sample sizes. Observe the reduction in variance at around 1M transactions.}
    \label{fig:ips-var}
\end{figure}

\begin{figure}[h]
    \centering
    \includegraphics[width=\textwidth]{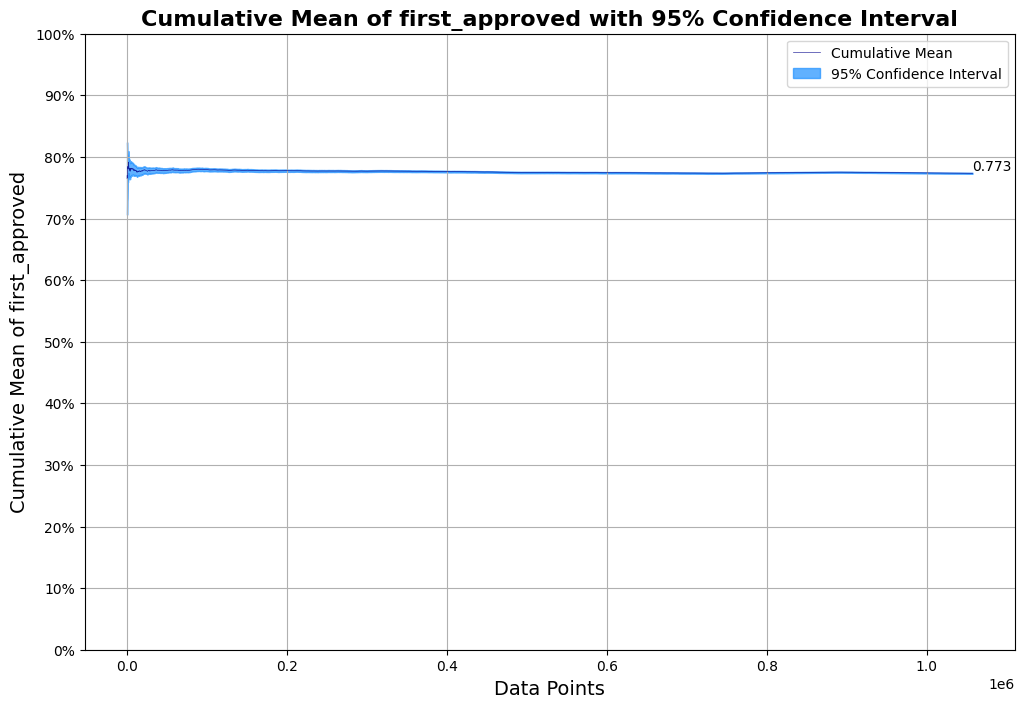} 
    \caption{Variance of On-policy Estimate across sample sizes. Observe the marked lack of variance compared to the IPS estimator in Figure \ref{fig:ips-var}.}
    \label{fig:ab-var}
\end{figure}

In contrast to the strong performance of IPS and SNIPS, DM consistently exhibits a negative correlation with online metrics. This suggests that the direct reward model struggles to accurately capture the underlying reward dynamics, leading to poor performance in off-policy evaluation. Similarly, the DR estimator, which combines DM and IPS, also shows weak correlation, hovering around zero. 

The overall outcome of these experiments is that it is indeed possible to, with over 80\% correlation, run AB tests offline. We estimate that this is recovering an incremental 9-54 million transactions over a six-month period by accelerating the discovery of winning variants by up to 12 months.

\section{Discussion}
\label{sec:discussion}

Our experience deploying OPE at Adyen has yielded several key insights:

\subsection{Scale and Variance}

Despite the well-known challenges of variance in importance sampling-based OPE estimators, particularly when there's a significant mismatch between logging and target policies, we observed surprisingly stable and reliable performance. This positive outcome can be largely attributed to scale. As seen in Figure \ref{fig:ips-var}, variance mostly disappears after 1M transactions---a scale that is not often seen in most benchmark datasets. Similar trends have been reported at Alibaba~\cite{zhou2018offline} and Netflix~\cite{gomez2021netflix}, reinforcing the idea that large-scale data can smooth out variance issues in OPE.

\subsection{OPE as a Complement to A/B Testing}

A critical insight from our deployment is that OPE estimates should be viewed as a prerequisite for on-policy A/B tests, rather than a complete replacement. While OPE provides rapid and cost-effective preliminary evaluations, false negatives (discarding a superior variant) can be more detrimental than false positives (overestimating a non-optimal variant). A false positive will be vetted and potentially rejected during a subsequent A/B test, whereas a false negative might permanently miss an opportunity to improve transaction performance. Therefore, we use OPE to filter and prioritize promising variants for further A/B testing, striking a balance between speed and thoroughness. This workflow mirrors industry practices at Criteo and Microsoft, where OPE is used to conduct early-stage filtering of suboptimal models~\cite{Gilotte_2018,bottou2013counterfactual,agarwal2016making}.

\subsection{Paradoxical Estimator Performance}

Our results highlight unexpected performance results of different OPE estimators. While IPS and SNIPS demonstrated strong and consistent correlation with online metrics, DM and DR exhibited weaker performance. A more expected outcome would be that IPS is the poor performer due to the high variance induced by the divergent policies. The poor performance of DM could also be expected and suggests that the direct reward model may be misspecified or inadequate for capturing the complex reward dynamics in our data. This, in turn, likely contributes to the underwhelming performance of DR, as it relies on the same underlying reward model. Further investigation is needed to improve the accuracy of the reward model and explore potential modifications to the DR estimator to enhance its effectiveness in our setting.

\section{Conclusion}
\label{sec:conclusion}

Off-Policy Evaluation offers a powerful alternative to traditional, purely on-policy evaluation methods for recommender systems, especially in data-intensive industries such as financial payments. Our deployment at Adyen shows that OPE can accelerate experimentation, reduce the costs of inconclusive A/B tests, and help identify winning strategies more rapidly.

By addressing practical challenges---such as variance control in importance sampling, leveraging exploration traffic, and scaling infrastructure---we demonstrated the strong alignment of OPE estimates with real-world A/B test outcomes. These findings provide a roadmap for other organizations seeking to adopt OPE in large-scale production settings. However, we emphasize that OPE is best used as a complementary tool alongside traditional on-policy evaluation. Together, these methods form a robust, multi-phase testing and validation pipeline that balances experimentation speed with business-critical reliability. Future work will be focused on taking OPE its logical conclusion and learning the reward directly E2E using Off-policy Learning techniques.  

\bibliographystyle{alpha}
\bibliography{vs}

\end{document}